\documentclass[conference]{IEEEtran}
\IEEEoverridecommandlockouts
\usepackage{cite}
\usepackage{bm}
\usepackage{amsmath}
\usepackage{graphicx}
\usepackage{textcomp}
\usepackage{xcolor}
\def\BibTeX{{\rm B\kern-.05em{\sc i\kern-.025em b}\kern-.08em
    T\kern-.1667em\lower.7ex\hbox{E}\kern-.125emX}}

\usepackage[T1]{fontenc} 
\usepackage[utf8]{inputenc}
\usepackage[american]{babel}
\usepackage[hidelinks]{hyperref}
\usepackage[capitalize]{cleveref}
\usepackage{mathtools}
\usepackage{booktabs}
\usepackage{fancyhdr}
\usepackage[shortcuts]{extdash}
\usepackage[all]{foreign}
\usepackage{bm}
\usepackage{tikz}
\usetikzlibrary{bayesnet}
\usepackage{siunitx}
\usepackage{xfrac}
\usepackage{amsfonts}

\usepackage{booktabs}
\usepackage{arydshln}
\usepackage{comment}
\usepackage{algorithm}
\usepackage{algpseudocode}
\usepackage{subcaption}

\usepackage[helvratio=0.92]{newtxtext}

\usepackage{temporal-logic}

\usepackage{glossaries}

\newacronym{ad}{AD}{Annotated Disjunction}
\newacronym{ann}{ANN}{Artificial Neural Network}
\newacronym{ap}{AP}{Access Point}
\newacronym{cfr}{CFR}{Channel Frequency Response}
\newacronym{cmis}{CMISymb}{Conditional Mutual Information on Symbolic data}
\newacronym{cnn}{CNN}{Convolutional Neural Network}
\newacronym{csi}{CSI}{Channel State Information}
\newacronym{cv}{CV}{Computer Vision}
\newacronym{dl}{DL}{Deep Learning}
\newacronym{edl}{EDL}{Evidential Deep Learning}
\newacronym{har}{HAR}{Human Activity Recognition}
\newacronym{kl}{\mbox{KL}}{Kullback–Leibler}
\newacronym{lan}{LAN}{Local-Area Network}
\newacronym{lif}{LIF}{Leaky Integrate-and-Fire}
\newacronym{lpcmci}{LPCMCI}{Latent PCMCI}
\newacronym{lstm}{LSTM}{Long Short-Term Memory}
\newacronym{ltl}{LTL}{Linear Temporal Logic}
\newacronym{mimo}{MIMO}{Multiple-Input Multiple-Output}
\newacronym{mlp}{MLP}{Multi-Layer Perceptron}
\newacronym{mse}{MSE}{Mean Squared Error}
\newacronym{nad}{nAD}{Neural Annotated Disjunction}
\newacronym{nic}{NIC}{Network Interface Card}
\newacronym{nn}{NN}{Neural Network}
\newacronym{ood}{OoD}{Out-of-Distribution}
\newacronym{ofdm}{OFDM}{Orthogonal Frequency-Division Multiplexing}
\newacronym{ofdma}{OFDMA}{Orthogonal Frequency-Division Multiple Access}
\newacronym{pcmci}{PCMCI}{Peter-Clark Momentary Conditional Independence}
\newacronym{phy}{PHY}{Physical Layer}
\newacronym{sdd}{SDD}{Sentential Decision Diagrams}
\newacronym{sdr}{SDR}{Software-Defined Radio}
\newacronym{siso}{SISO}{Single-Input Single-Output}
\newacronym{snn}{SNN}{Spiking Neural Network}
\newacronym{std}{STD}{Standard Deviation}
\newacronym{stdp}{STDP}{Spike-Timing-Dependent Plasticity}
\newacronym{sta}{STA}{station}
\newacronym{vae}{VAE}{Variational Auto-Encoder}
\newacronym{wmc}{WMC}{Weighted Model Counting}
\newacronym{wlan}{WLAN}{wireless Local-Area Network}

\newacronym{mdl}{MDL}{Minimum Description Length}
\newacronym{leace}{LEACE}{Least-Squares Concept Erasure}
\newacronym{dft}{DFT}{Discrete Fourier Transform}
\newacronym{tpe}{TPE}{Tree-structured Parzen Estimator}
\newacronym{rmse}{RMSE}{Root Mean Squared Error}

\usepackage{amsthm}

\usepackage{amssymb}  

\usepackage{mathtools}

\begin{document}

\title{Preliminary Insights in Chronos Frequency Data Understanding and Reconstruction}

\author{
    \IEEEauthorblockN{
        Alessandro Pagani\IEEEauthorrefmark{1},
        Marco Cominelli\IEEEauthorrefmark{3},
        Liying Han\IEEEauthorrefmark{5},
        Gaofeng Dong\IEEEauthorrefmark{5},
        Sergio Benini\IEEEauthorrefmark{1},\\
        Francesco Gringoli\IEEEauthorrefmark{1},
        Mattia Savardi\IEEEauthorrefmark{1},
        Mani B. Srivastava\IEEEauthorrefmark{5},
        Trevor Bihl\IEEEauthorrefmark{6},
        Erik P. Blasch\IEEEauthorrefmark{7},\\
        Daniel O. Brigham\IEEEauthorrefmark{7},
        Kara Combs\IEEEauthorrefmark{7},
        Lance M. Kaplan\IEEEauthorrefmark{4}, 
        and Federico Cerutti\IEEEauthorrefmark{1}\IEEEauthorrefmark{8}\IEEEauthorrefmark{9}\IEEEauthorrefmark{10}
    }\\
    \IEEEauthorblockA{
        \IEEEauthorrefmark{1}%
            DII, University of Brescia, Italy. \\
            \{alessandro.pagani, sergio.benini, francesco.gringoli, mattia.savardi federico.cerutti\}@unibs.it
    }
    \IEEEauthorblockA{
        \IEEEauthorrefmark{3}%
            DEIB, 
            Politecnico di Milano, Italy. 
            marco.cominelli@polimi.it
    }
    \IEEEauthorblockA{
        \IEEEauthorrefmark{5}%
            ECE Department,
            University of California, Los Angeles, USA.
            \{liying98,gfdong,mbs\}@ucla.edu
    }
    \IEEEauthorblockA{
        \IEEEauthorrefmark{6}%
            Ohio University, USA.
            bihlt@ohio.edu
    }
    \IEEEauthorblockA{
        \IEEEauthorrefmark{7}%
            Air Force Research Laboratory, USA. 
            \{erik.blasch.1, daniel.brigham, kara.combs.1\}@us.af.mil
    }
    \IEEEauthorblockA{
        \IEEEauthorrefmark{4}%
            DEVCOM Army Research Lab, USA.
            lance.m.kaplan.civ@army.mil
    }
    \IEEEauthorblockA{
        \IEEEauthorrefmark{8}%
            Cardiff University, UK.
    }
    \IEEEauthorblockA{
        \IEEEauthorrefmark{9}%
            University of Southampton, UK.
    }
    \IEEEauthorblockA{
        \IEEEauthorrefmark{10}%
            Imperial College London, UK.
    }
}

\maketitle

\begin{abstract}
This paper presents a preliminary analysis of the ability of Chronos foundation model to process and internally represent frequency domain information. Foundation models that process time-series data offer practitioners a unified architecture capable of learning generic temporal representations across diverse tasks and domains, reducing the need for task-specific feature engineering and enabling transfer across signal modalities. Despite their growing adoption, the extent to which such models encode fundamental signal properties remains insufficiently characterised. We address this gap by analysing Chronos under controlled conditions, starting from the simplest class of signals: discrete sinusoids generated at fixed frequencies. Using lightweight online minimum description length probes applied to the decoder architecture, we test for the presence and separability of frequency information in the model’s internal representations. The results provide insight into how frequential content is captured across the frequency spectrum and highlight regimes in which representation quality may degrade or require particular care. These findings offer practical guidance for users of Chronos in signal processing and information fusion contexts, and contribute to ongoing efforts to improve the interpretability and evaluation of foundation models for temporal data.
\end{abstract}


\section{Introduction}
\label{sec:introduction}
Foundation models for temporal data aim to learn generic representations that transfer across forecasting, classification, and generative tasks without task-specific feature engineering. Amazon’s Chronos \cite{ansari2024chronoslearninglanguagetime} belongs to this class of pre-trained sequence models and has been adopted in a range of signal processing contexts \cite{11180285, park2025unicastunifiedmultimodalprompting, olivares2025realisticevaluationcrossfrequencytransfer, PerakylaYear}. Despite their empirical success, a basic question remains insufficiently characterised: to what extent do such models encode fundamental signal properties in their internal representations?

In classical signal processing, frequency constitutes one of the most elementary and interpretable attributes of a time series. The ability to represent and manipulate frequential content underpins forecasting, filtering, detection, and information fusion pipelines \cite{10706320}. If a temporal foundation model is to serve as a general-purpose representation learner, frequency information should be explicitly and reliably encoded in its latent states \cite{11123929}. However, end-to-end performance on downstream tasks does not by itself establish whether such information is linearly accessible, redundantly distributed, or fragile under architectural constraints.

While various iterations of the Amazon Chronos model exist \--- including the latest Chronos-2 \--- this work deliberately employs Chronos-Bolt\footnote{We used the ``chronos-bolt-tiny'' variant. Hugging Face model ID: \href{https://huggingface.co/amazon/chronos-bolt-tiny}{\texttt{amazon/chronos-bolt-tiny}} (accessed February 9, 2026).} \cite{ansari2025chronos2univariateuniversalforecasting, ansari2024chronoslearninglanguagetime, chronos_github} to ensure a clean and controlled baseline. To further simplify the experimental setting, we restrict our evaluation to the simplest non-trivial signal class: discrete sinusoids generated at fixed frequencies within a limited operational band. Under this setup, we first evaluate the separability of the frequency discrimination task, identifying which frequencies are most affected. We then investigate causality by removing the corresponding information and measuring the resulting degradation. With Chronos kept frozen throughout, the spectrum is recursively partitioned into hierarchical binary bands, enabling progressively finer discrimination tasks and allowing explicit identification of degradation near decision boundaries. We extract decoder hidden states and apply lightweight online \gls{mdl} probes \cite{bornschein2023sequential, voita-titov-2020-information} to assess whether frequency-related information is present and linearly accessible. 
Probe performance is evaluated using information-theoretic compression metrics relative to a non-informative baseline, providing a quantitative measure of how strongly the model's internal states reflect frequency structure.

Beyond static probing, we perform an intervention study using \gls{leace} \cite{10.5555/3666122.3669006}. Linear subspaces associated with a selected binary frequency abstraction are removed sequentially across decoder layers, and the intervened model is evaluated in closed-loop autoregressive generation. 

This analysis contributes to the interpretability and evaluation of temporal foundation models in two primary respects:
\begin{enumerate}
    \item It establishes whether frequency information is explicitly and linearly separable in decoder representations, while also characterising structured degradation regimes associated with architectural synchronisation effects, such as patch-stride aliasing.
    \item It quantifies the redundancy and distribution of this frequency encoding across different model layers using erasure techniques to assess possible causal links.
\end{enumerate}
Ultimately, these insights offer practical guidance for deploying Chronos in signal processing and information fusion contexts where frequential fidelity is critical.

The rest of the paper is as follows: Sect. \ref{sec:background} provides a background for the Chronos-Bolt time series analysis, Sect. \ref{sec:methodology} list the experimental design, Sect. \ref{sec:results} results, and Sect. \ref{sec:conclusions} conclusions.

\section{Background}
\label{sec:background}
\subsection{The Chronos-Bolt Architecture}
Chronos-Bolt is a foundation model for time-series forecasting built on the T5 encoder--decoder architecture \cite{raffel2020exploring}, adapted to univariate sequence prediction.
In contrast to classical autoregressive models that predict one time step at a time, Chronos-Bolt processes contiguous patches of the input sequence and directly regresses multiple quantiles. Patch-based tokenisation reduces effective sequence length by grouping neighbouring samples, improving computational efficiency, while multi-quantile regression yields a direct approximation of the predictive distribution.

\paragraph{Generation} Given an initial context window $X_{1:T}$, the model generates a sequence of length $T= 512$ in an autoregressive manner. At each step, it predicts (the default value of) $O=64$ future time steps, which are appended to the context $X$ before the next prediction.

\paragraph{Preprocessing and Tokenisation}

Given an input time-series data $X_{1:T}$, instance standard score normalization is applied:
\begin{equation}
    \tilde{X}_t = \frac{X_t - \mu} {\max(\sigma, \epsilon)}, \text{ with } t=1,\ldots,T
    \label{eq:instance_norm}
\end{equation}
where $\mu$ and $\sigma$ are the empirical mean and standard deviation computed over the context window using a population-style estimator, with $\epsilon$ ensuring numerical stability. The normalised sequence is partitioned into non-overlapping patches of fixed length $W$ and equal stride. Each patch $\mathbf{x}$ is projected into the transformer hidden dimension $d_{\text{model}}$ through a residual block:
\begin{equation}
\begin{aligned}
    \mathbf{m} &= \text{Sigmoid}(\mathbf{W}_m \mathbf{x} + \mathbf{b}_m), \\
    \mathbf{o} &= \text{Dropout}(\mathbf{W}_o \mathbf{h} + \mathbf{b}_o), \\
    \mathbf{r} &= \mathbf{W}_r \mathbf{x} + \mathbf{b}_r, \\
    \mathbf{m'} &= \text{LayerNorm}(\mathbf{o} + \mathbf{r}),
\end{aligned}
\end{equation}
where $\mathbf{W}$ are the weights, $b$ the biases, and $\mathbf{r}$ provides the residual connection. 
The resulting embeddings form the encoder input sequence.

\paragraph{Transformer Backbone}

The backbone follows the standard T5 stack \cite{10.5555/3455716.3455856}. The encoder comprises repeated layers of multi-head self-attention and position-wise feed-forward sublayers, each wrapped with pre-normalisation and residual connections. The decoder mirrors this structure but introduces masked self-attention and encoder--decoder cross-attention. Masking enforces causality within the prediction window, while cross-attention conditions forecasts on the full encoded context. This transformer architecture yields contextualised hidden states at each decoder layer, which we later analyse using probing methods.

\paragraph{Direct Quantile Prediction}

The decoder’s final hidden states are mapped, via an output residual block, from $d_{\text{model}}$ to a fixed set of quantiles over the prediction horizon $O$. During training, future targets are normalised using the same $(\mu, \sigma)$ computed from the input context (similar to Equation~\ref{eq:instance_norm}).

\subsection{Probing Classifiers and Causality}
\label{sec:background:probeMDL}
\subsubsection{Probing with Minimum Description Length}
Probing Classifiers assess properties encoded in frozen neural representations $h$ by predicting targets $y$ \cite{hewitt-liang-2019-designing, voita-titov-2020-information}. To evaluate representational quality through compression, we employ sequential Minimum Description Length (\gls{mdl}) estimation \cite{bornschein2023sequential}. For a sequence of $K$ batches $\mathcal{B} = (B_1, \dots, B_K)$, the total prequential codelength is the cumulative next-step log-loss:
\[
L_{total} = - \sum_{k=1}^{K} \sum_{(h, y) \in B_k} \log_2 p_{\theta_{k-1}}(y \mid h)
\]
Crucially, the probe parameters $\theta_{k-1}$ are evaluated on the unseen $k$-th batch before being updated. To efficiently approximate training over the full past history without large in-memory buffers, we update the probe online using replay streams, which augment the immediate training data with older past samples. Under this framework, a shorter total codelength indicates that the target property is efficiently encoded within the representation.

To evaluate probe selectivity, we adopt control tasks in which the true labels \( y \) are replaced with uniformly sampled random labels \( y_{\text{control}} \), while keeping all other settings identical \cite{hewitt-liang-2019-designing, voita-titov-2020-information}. Substantial compression or high accuracy in this setting would indicate memorisation rather than genuine extractability. A selective probe should reduce codelength on the true task but fail to compress the control task, yielding a description length comparable to or exceeding the uniform reference.

\subsubsection{Searching for causal evidence}

Probing establishes correlational evidence that a concept is linearly decodable from a representation, but does not by itself establish causal relevance. To assess causal contribution, we apply \gls{leace} \cite{10.5555/3666122.3669006}, which removes linear information about a target concept.

Let $h \in \mathbb{R}^d$ denote representations and $y \in \mathbb{R}^k$ the concept. By keeping $k < d$, the null space of $\Sigma_{hy}$ exists to compose the matrix P at least with dimension $d - k$. This allows $P$ to be an orthogonal projection matrix that ``cleans'' the concept dimensions while preserving the remaining $d - k$ dimensions of the representation. \gls{leace} seeks an affine transformation $\psi(h) = Ph + b$ such that the transformed representations are linearly guarded against $y$, i.e.\ $\Sigma_{hy} = \mathrm{Cov}(h,y)$ satisfies
\begin{equation}
    P \Sigma_{hy} = 0.
\end{equation}
Among all such transformations, \gls{leace} minimises distortions:
\begin{equation}
    \min_{P,b} \ \mathbb{E}\!\left[\|h - (Ph + b)\|^2\right] 
    \quad \text{s.t.} \quad \mathrm{Cov}(Ph+b, y) = 0.
\end{equation}
The resulting closed-form solution projects $h$ onto the nullspace of $\Sigma_{hy}$ (after whitening) and then reverses the whitening transformation, ensuring exact linear erasure with minimal alteration of unrelated structure. Once computed, the transformation is applied at inference time without further optimisation.

\section{Methodology}
\label{sec:methodology}
\begin{figure}[!t]
  \centering
  \includegraphics[width=2.2in] {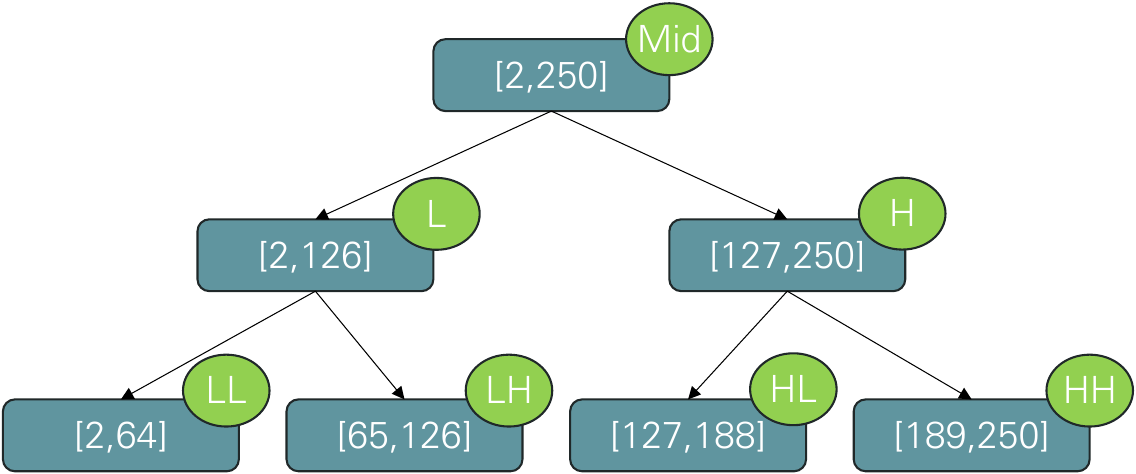}
  \caption{Hierarchical frequency splitting strategy. The spectrum is divided into seven intervals through iterative binary partitioning; green circular markers denote the specific test assignments for each frequency segment.}
  \label{fig:methodology:1}
\end{figure}

\subsection{Implication of the Nyquist-Shannon Sampling Theorem}
For our evaluations, the input context window for Chronos-Bolt was restricted to 512 tokens. This length was specifically chosen to ensure direct comparability across the entire Chronos model family, as the original Chronos-T5 architecture requires a strict 512-token input limit. For simplicity of treatment, unless specified otherwise, we assume, in the following, that this corresponds to a single second of a signal sampled at $512$ Hz.
We thus generate synthetic datasets composed of sinusoids, \ie
\begin{equation}
    X_f[n] = \sin(2\pi f n / f_s)
\end{equation}
setting the sampling frequency to $f_s = 512$ Hz and the length of the sinusoids to $T=512$ (samples). Consequently, we restrict the frequency $f$ to the interval $[2,250]$ Hz. The upper bound is chosen to strictly satisfy the Nyquist-Shannon sampling theorem: given the sampling rate, the cutoff at 250 Hz provides a necessary guard band below the Nyquist frequency ($f_s/2 = 256$ Hz) to prevent aliasing artifacts. The lower bound of 2 Hz is established to ensure that all generated signals exhibit sufficient periodicity within the observation window.

We employ spectral predictability~$\Omega$ \cite{wang2025spectral} as a measure of intrinsic forecastability and as a model selection criterion. Large-scale evaluations show that time series foundation models outperform lightweight baselines when $\Omega$ is high, with the performance gap narrowing as $\Omega$ decreases. Since $\Omega$ can be computed within seconds per dataset, it offers an efficient reliability indicator without extensive validation.

Following \cite{wang2025spectral}, each time series is transformed via the Fast Fourier Transform (FFT) into a power spectrum, and normalised to form a probability distribution. Its Shannon entropy is computed, and predictability is defined as the inverted, normalised entropy, yielding $\Omega \in [0,1]$. High $\Omega$ indicates concentrated spectral mass and strong periodic structure; low $\Omega$ reflects diffuse, noise-like behaviour.

Because our dataset consists of clean sinusoids, it exhibits high $\Omega$, supporting the use of Chronos for forecasting.

\subsection{Probing}
\label{sec:task_extr}

We cast the analysis as a \textit{probing task} to determine whether Chronos’ decoder representations encode frequency information. The operational band \( [2, 250] \) Hz is recursively partitioned into seven bands and sub-bands (Figure~\ref{fig:methodology:1}), enabling controlled evaluation across progressively finer intervals and quantification of aleatory uncertainty near decision boundaries.

To balance coverage across frequencies, sampling is governed by the number of distinct phase shifts available per sinusoid. For frequency \( f \), the maximum number of unique shifts is
\begin{equation}
S_f = \min \left\{\frac{f_s}{\gcd\left( f, f_s \right)} - 1, N \right\},
\label{eq:n_splits}
\end{equation}
where $\gcd(a,b)$ denotes the greatest common divisor of $a$ and $b$. This retains only non-redundant phase configurations while capping the sample count at \( N \).

For each \( f \), discrete sinusoids are generated and segmented with a sliding window of length \( T = 512 \) and stride 1, producing \( S_f \) distinct segments \( \mathbf{x} \). Windows are randomly split into training, validation, and test sets with an explicit uniqueness check to prevent duplication across splits.

\subsubsection{Probing with MDL}
\begin{figure}[!t]
    \centering
    \includegraphics[width=2in]{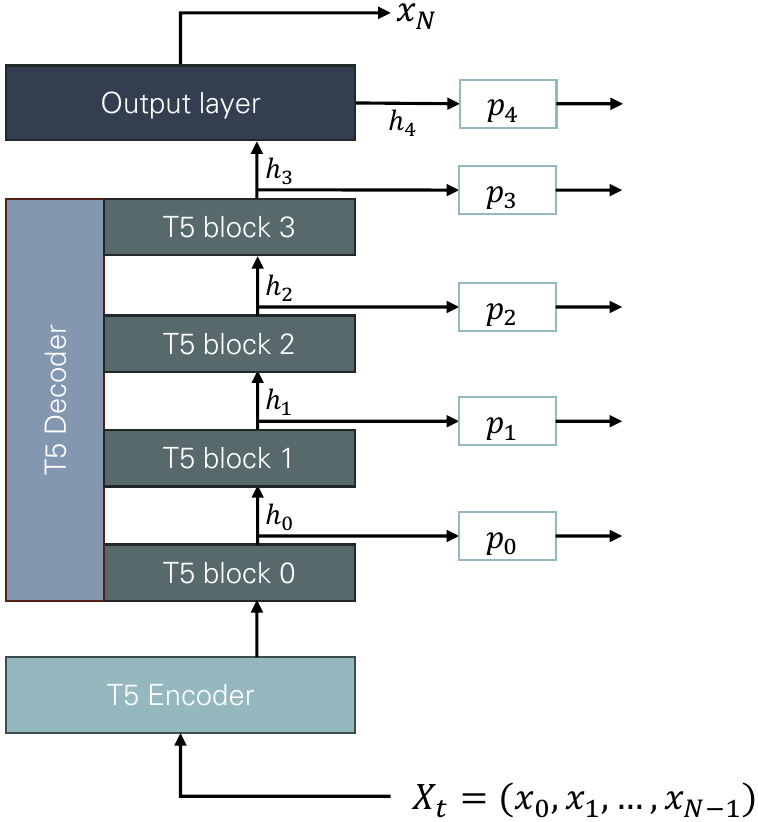}
    \caption{Overview of the Chronos architecture, with a specific focus on the decoder module. The diagram illustrates the placement of linear probes at five distinct stages: immediately following decoder blocks 0, 1, 2, and 3, as well as after the final output layer.}
    \label{fig:methodology:2}
\end{figure}

Segments are passed through Chronos with frozen parameters, and decoder hidden states are extracted (Figure~\ref{fig:methodology:2}). Lightweight online \gls{mdl} probes are trained on these representations to predict either frequency identity or binary band labels. The resulting codelength quantifies how efficiently frequency information can be recovered from the representations.

The training of the probing classifiers is performed employing the Lightweight \gls{mdl} described in Section ~\ref{sec:background:probeMDL}. Moreover, since the raw codelength is intrinsically dependent on the sample size of the training set, comparing raw values across different data splits can be misleading. To obtain a normalised measure of extractable information, we compute the \emph{Space Saving} ($SV$) associated with each probing task. Let \( L(D) \) denote the total codelength required to transmit the dataset labels for certain test using the trained \gls{mdl} probe, \ie the cumulative description length after the online coding procedure. Let \( L_{\text{uniform}}(D) \) denote the reference codelength obtained under a naive uniform prior over labels, corresponding to random guessing (for binary tasks, a fair coin).

The Space Saving \--- adapted from \cite{IJISTECH242} \--- is defined as
\begin{equation}
    SV = 1 - \frac{L(D)}{L_{\text{uniform}}(D)}.
    \label{eq:compression}
\end{equation}

This quantity measures the relative reduction in coding cost achieved by the probe compared to the uninformed baseline. Values \( S \rightarrow 1 \) indicate that the representations render the labels highly compressible, corresponding to strong linear separability. Conversely, \( S \rightarrow 0 \) implies that the probe offers no compression advantage over the uniform prior, indicating that the representations do not encode exploitable information about the target labels \cite{voita-titov-2020-information}.

To characterise how frequency information varies across the spectrum, we evaluate probe performance at the individual frequency level. Although $SV$ is the primary metric, classification accuracy provides a complementary and interpretable measure; since accuracy and codelength are not strictly equivalent \cite{voita-titov-2020-information}, we first verify their empirical alignment. Using the test set from Section~\ref{sec:task_extr}, samples are grouped by true frequency and accuracy is computed per frequency, yielding a spectral performance profile over the operational band. This stratification reveals the degradation gap, \ie the boundary region where performance transitions from reliable discrimination to chance level, thereby complementing global compression metrics with local analysis.

\subsubsection{Searching for causal evidence}

Linear separability provides an operational indicator of how task-relevant information is structured within the learned representations. To move beyond correlation, we introduce an \textit{intervention} to identify a possible causal mechanism: we remove the linear subspace associated with a given binary task and evaluate the resulting impact on spectral performance. Although this procedure can be applied to any task in the spectral hierarchy, we focus on ``Task Mid'', which spans the full operational frequency range. This allows us to assess whether removing its associated subspace produces localised degradation or broader performance collapse.

For this analysis, we adopt a continuous data generation scheme rather than the discrete construction described in Section~\ref{sec:task_extr}. For each frequency \( f \), we generate a sinusoid of length \( 2T \) defined as
\[
X'_f[n] = \sin\left( \frac{2\pi f n}{f_s} + \phi_i \right),
\]
where $\phi_i \sim \text{Uniform}([0, 2\pi))$ and $n = 0, \dots, 2T-1$.
A sliding window of length $T=512$ and stride 1 is applied to obtain time series segments. The total number of windows equals the number of sampled phases (100). These segments are randomly split into training and test sets.

To remove task-specific information from the hidden states, we apply \gls{leace} \cite{10.5555/3666122.3669006}, which computes a closed-form linear transformation that eliminates the target concept—here, the binary task labels—while minimising \( L_2 \) distortion of the original representation. After intervention, the task is provably linearly unlearnable from the modified features. The procedure is implemented using the \texttt{concept-erasure} library \cite{concepterasure_github}.

\begin{algorithm}[!t]
\caption{Sequential LEACE Eraser Fitting}
\label{alg:sequential_eraser}
\begin{algorithmic}[1]
\Require Pre-trained model $\mathcal{M}$, ordered target layers $L = \{l_1, l_2, \dots, l_k\}$, training inputs $X$, concept labels $y$
\Ensure Fitted erasers $\mathcal{E} = \{E_1, E_2, \dots, E_k\}$
\State $\mathcal{E}_{active} \gets \emptyset$
\For{$i = 1$ \textbf{to} $k$}
    \State $\mathbf{h}_{l_i} \gets \textsc{ForwardPass}(\mathcal{M}, X, \mathcal{E}_{active})$
    \State $E_i \gets \textsc{FitLEACE}(\mathbf{h}_{l_i}, y)$
    \State $\textsc{Freeze}(E_i)$
    \State $\mathcal{E}_{active} \gets \mathcal{E}_{active} \cup \{E_i\}$
\EndFor
\State \Return $\mathcal{E}_{active}$
\end{algorithmic}
\end{algorithm}

Algorithm~\ref{alg:sequential_eraser} describes the sequential intervention across selected layers. We initialise an empty set of active erasers and iterate through layers in forward order. At each layer $l_i$, a forward pass is performed with all previously fitted erasers applied, ensuring that the extracted activations $\mathbf{h}_{l_i}$ reflect the already-intervened network state. A  projector $E_i$ is then fitted to remove the linear dependence between $\mathbf{h}_{l_i}$ and the target concept $y$. The fitted eraser is frozen and added to the active set so that it remains applied in subsequent layers and during evaluation. This sequential procedure ensures that each $E_i$ removes residual concept information not eliminated by earlier interventions.

After fitting $\{E_1, \dots, E_k\}$, all erasers are activated and the intervened model is evaluated using a closed-loop autoregressive generation task.

\paragraph{Generation} Given an initial context window $\mathbf{x}_{0:T-1}$, the model generates a sequence $\hat{\mathbf{x}}_{gen}$ of length $T = 512$ in an autoregressive manner. At each step, it predicts 64 future time steps, which are appended to the context $X$ before the next prediction. The forecasting horizon of 64 follows the default Chronos configuration.

\paragraph{Spectral analysis} For each generated sequence $\hat{\mathbf{x}}_{gen}$, we compute its \gls{dft}. The dominant frequency is estimated as the index of the maximum magnitude component:
\[
\hat{f} = \arg\max_{k} 
\left|
\sum_{n=0}^{T-1} 
\hat{x}_{gen}[n] 
e^{-i 2\pi k n / T}
\right|.
\]

\paragraph{Metric} We quantify frequency degradation using the mean squared error between the ground-truth input frequency $f_i$ and the dominant generated frequency $\hat{f}_i$:
\[
MSE = \frac{1}{n} \sum_{i=1}^{n} (f_i - \hat{f}_i)^2,
\]
where $f_i$ denotes the true input frequency and $\hat{f}_i$ the frequency estimated from the generated sequence.

\subsection{Implementation Details}

\begin{table}[tbp]
\centering
\footnotesize
\caption{Hyperparameter search space and sampling distributions.}
\label{tab:hyperparameters}
\begin{tabular}{lll}
\toprule
\textbf{Hyperparameter} & \textbf{Search Space} & \textbf{Distribution} \\
\midrule
\multicolumn{3}{@{}l}{\textit{Lightweight MDL}} \\
Replay Streams & $[1, 5]$ & Uniform (Int.) \\
EMA Decay & $[0.005, 0.1]$ & Log-Uniform \\
Reset Prob. & $[0.01, 0.2]$ & Uniform \\
Noise Level & $[0.01, 0.1]$ & Uniform \\
\midrule
\multicolumn{3}{@{}l}{\textit{Probe}} \\
Batch Size & $\{64, 128, 256\}$ & Categorical \\
Learning Rate & $[10^{-5}, 10^{-1}]$ & Log-Uniform \\
Weight Decay & $[10^{-5}, 10^{-2}]$ & Log-Uniform \\
Dropout & $[0.1, 0.3]$ & Uniform \\
\bottomrule
\end{tabular}
\end{table}

All models were implemented in PyTorch~\cite{NEURIPS2019_bdbca288}. Hyperparameters were optimised using Optuna \cite{10.1145/3292500.3330701} with 100 trials and the \gls{tpe}, minimising total codelength. The explored ranges are reported in Table~\ref{tab:hyperparameters}. Log-uniform sampling was used for scale-sensitive parameters to ensure coverage across orders of magnitude. The prequential \gls{mdl} implementation follows the sequential learning framework of \cite{bornschein2023sequential}, adapted to our spectral probing setting while preserving the online continual learning and replay-stream mechanisms. \gls{leace} was applied with default settings, and successful concept removal was verified during training. We used the median value produced by chronos-bolt-tiny model to evaluate the FFTs. 

\section{Results}
\label{sec:results}

\begin{figure}[!t]
    \centering
    \includegraphics[width=3.5in]{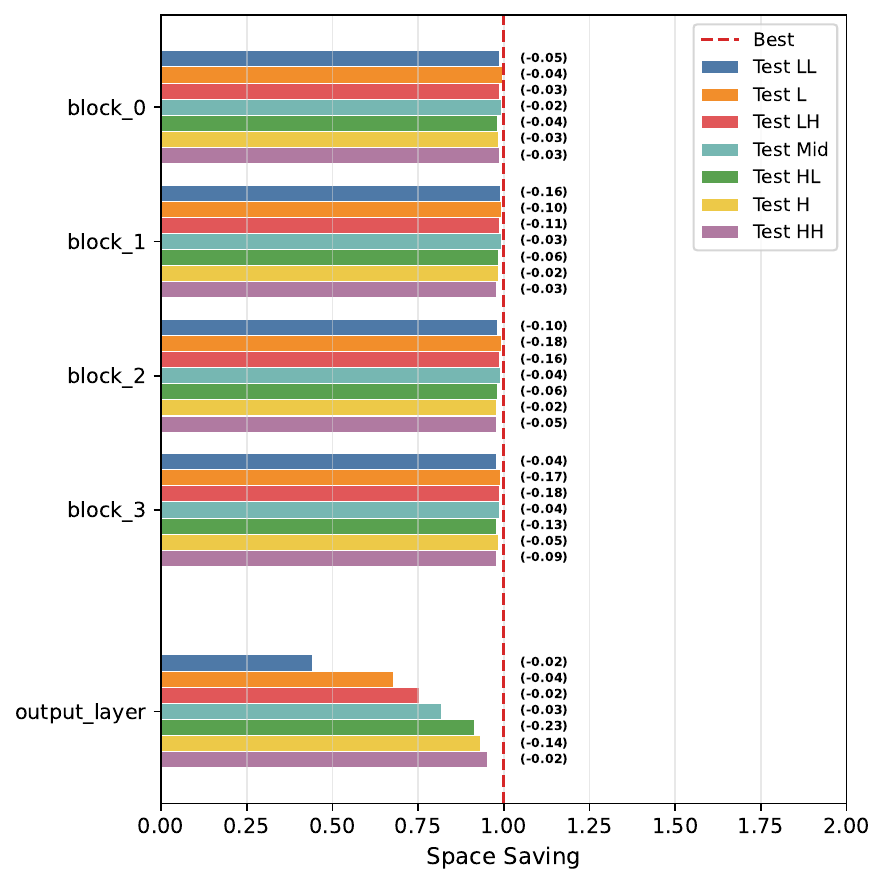}
    \caption{Horizontal grouped bar chart illustrating compression performance across the network architecture. The y-axis groups results into five structural stages (four macro blocks and the final output layer). Within each stage, seven bars represent individual tests, ordered by ascending classification threshold. Bar lengths represent the $SV$ for the target task, while values in parentheses indicate the corresponding $SV$ obtained on the control task.}
    \label{fig:results:1}
\end{figure}

\begin{figure*}[!t]
    \centering
    \includegraphics[width=6in]{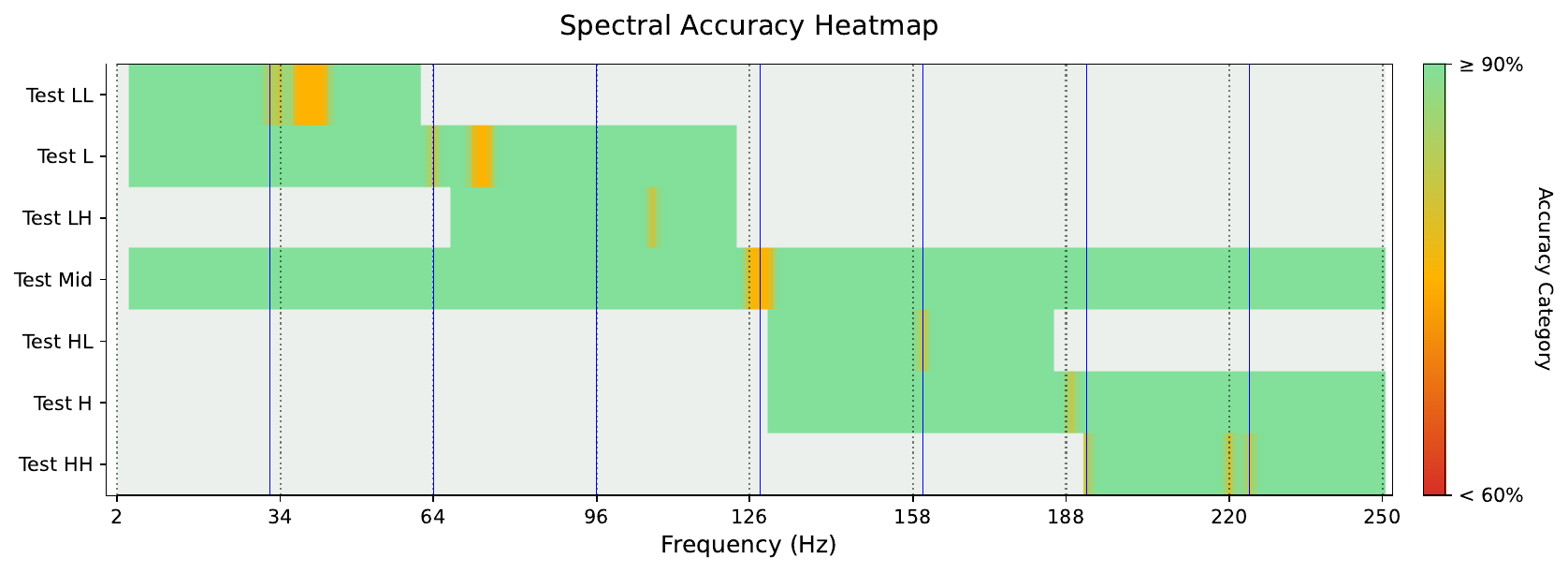}
    \caption{Task-stratified spectral accuracy heatmap. The visualization displays classification performance across frequency bins (Hz) for seven ordered tasks (Test LL–Test HH). Color indicates thresholded accuracy levels ranging from a failure to generalize ($<$ 0.6, Red) and optimal performance ($\geq$ 0.9, Green). Grey regions indicate frequency bands excluded from the analysis. Blue thin vertical lines corresponds to 32 Hz frequency multiples from 1 to 7.}
    \label{fig:results:3}
\end{figure*}

\subsection{MDL Results}
Our analysis highlights two key findings, which we explore below: internal decoder layers preserve frequency features much better than the final output and the presence of patch-induced accuracy dependencies.

Figure~\ref{fig:results:1} reports $SV$ results (Equation~\ref{eq:compression}). The four macro decoder blocks exhibit a flat profile with compression ratios approaching 1, indicating that task-relevant frequency information is linearly accessible at these stages. In contrast, the output layer displays a clear spectral dependence: lower input frequencies yield higher codelengths and thus reduced compression. This increased description length reflects a greater linear complexity required for classification, implying diminished extractability at the final stage. Overall, \textit{frequency information is more robustly encoded within the internal decoder representations} than at the output layer.

For the control task (Figure~\ref{fig:results:1}, in parentheses), which uses random labels to test if the network overlearns, the codelengths exceed the uniform baseline ($SV \leq 0$). This confirms that no genuine structure can be extracted from random labels. The excess codelength corresponds to the overhead incurred when the probe attempts to model noise, demonstrating that the \gls{mdl} metric does not spuriously attribute structure to random signals.

\newcommand{\ra}[1]{\renewcommand{\arraystretch}{#1}}

The classification accuracy across the network closely follows its $SV$ profile. The first four decoder blocks maintain near-perfect accuracy, while the output layer's accuracy drops to a mean of 0.9567. Notably, these output values increase monotonically from 0.8605 to 0.9952, mirroring the overall trend of the $SV$ metric.

To move beyond aggregate metrics, we compute per-frequency test accuracy, allowing identification of spectral regions that are reliably separable versus those that collapse to chance. Figure~\ref{fig:results:3} reveals a consistent pattern. Except for \textit{Test LH}, all tasks show a local degradation in accuracy near the binary decision boundary. For \textit{LL}, \textit{Mid}, and \textit{H}, the drop persists for frequencies immediately above the threshold, indicating asymmetric separability.

A further observation is the presence of isolated frequency-specific failures across tasks, notably 32\,Hz in \textit{Test LL}, 112\,Hz in \textit{Test LH}, and 160/224\,Hz in \textit{Test HH}. These singular dips are unlikely to reflect task difficulty. Instead, they suggest structural artefacts, plausibly linked to discretisation effects induced by the Chronos-Bolt patching scheme.

Specifically, Chronos-Bolt-Tiny uses non-overlapping patches of length $P=16$ and stride $S=16$. 
With sampling frequency $f_s = 512$ Hz, the temporal duration of one patch is
\begin{equation}
\Delta T_p = \frac{P}{f_s} = \frac{16}{512} = 0.03125 \ \text{s}.
\end{equation}

Because of this fixed window, performance degrades when the input frequency becomes phase-locked to the patch grid. 
This occurs when the signal frequency is an integer multiple of the fundamental patch frequency
\begin{equation}
f = k f_p = k \left( \frac{f_s}{P} \right), 
\quad k \in \mathbb{Z}^{+}.
\end{equation}

Since $f_p = 32$ Hz, $k=1$ yields $f=32$ Hz, where exactly one full cycle fits inside a patch. 
For higher harmonics ($k \ge 2$), multiple cycles fall within each patch while preserving a fixed phase relation across tokens.

To formalize this relationship, let the sinusoid be
\begin{equation}
x[n] = \sin\!\left(2\pi \frac{f}{f_s} n + \phi \right),
\end{equation}
with patch index $j$ and $n \in [jP, (j+1)P-1]$. 
The $j$-th patch is
\begin{equation}
P_j = \{ x[n] \mid n \in [jP, (j+1)P-1] \}.
\end{equation}

Under the phase-locking condition $f = k(f_s/P)$,
\begin{equation}
\begin{aligned}
x[n+P] 
&= \sin\!\left(2\pi \frac{k}{P}(n+P) + \phi \right) \\
&= \sin\!\left(2\pi \frac{k}{P} n + 2\pi k + \phi \right) \\
&= x[n].
\end{aligned}
\end{equation}

Hence, successive patches are identical, $P_j = P_{j+1}$, producing a degenerate token sequence and reduced temporal resolution. 
This effect explains the sharp accuracy drops observed at harmonics of 32 Hz in several tasks. 

The behaviour is not strictly deterministic: some harmonics (e.g.,\ 96 Hz) do not degrade, while non-integer sub-harmonics (e.g.,\ 112 Hz) may exhibit failures. 
Patch-stride aliasing is therefore a primary mechanism, but positional encoding and non-linear interactions likely modulate its impact.

\begin{table}[tbp]
\centering
\ra{1.2}
\begin{tabular}{
l
S[table-format=5.2]
S[table-format=1.2e-1]
c
}
\toprule
\bfseries Layers affected 
& {\bfseries RMSE} 
& {\bfseries p-value} 
& {\bfseries Sig.} \\
\midrule
\textit{Baseline} & 137.71 & {} & {} \\
\midrule
Test 0     & 137.55 & 1.18e-02 & * \\
Test 1     & 137.98 & 5.65e-01 &  \\
Test 2     & 138.57 & 5.86e-05 & * \\
Test 3     & 139.92 & 3.93e-07 & * \\
Test 4     & 134.25 & 8.77e-08 & * \\
Test 01    & 138.20 & 7.90e-02 &  \\
Test 012   & 139.48 & 2.88e-06 & * \\
Test 0123  & 139.04 & 2.04e-05 & * \\
Test 01234 & 140.11 & 7.06e-09 & * \\
\textbf{Test 1234} & \bfseries 140.75 & 4.00e-10 & * \\
Test 234   & 139.59 & 3.44e-06 & * \\
Test 34    & 140.62 & 3.15e-08 & * \\
\bottomrule
\end{tabular}
\caption{Frequency concept erasure results. Spectral \gls{rmse} is reported for the baseline Chronos model and each sequential \gls{leace} intervention (block indices indicate affected layers; block 4 corresponds to the output layer). P-values are obtained from two-sided paired Wilcoxon signed-rank tests against baseline. Asterisks denote statistically significant differences at $\alpha=0.05$. Higher \gls{rmse} indicates stronger spectral degradation and thus more effective concept erasure.}
\label{table_frequency_degr_merged}
\end{table}

\subsection{Searching for Causal Evidence}
We assess sequential \gls{leace} by erasing the binary abstraction defined by \textit{Task Mid} (low vs.\ high frequency) rather than continuous frequency values. This tests whether removing a coarse categorical representation suffices to impair fine-grained spectral generation. Performance is measured via Spectral \gls{rmse}, where higher values indicate stronger degradation.

\begin{figure}[!t]
    \centering
    \includegraphics[width=3in]{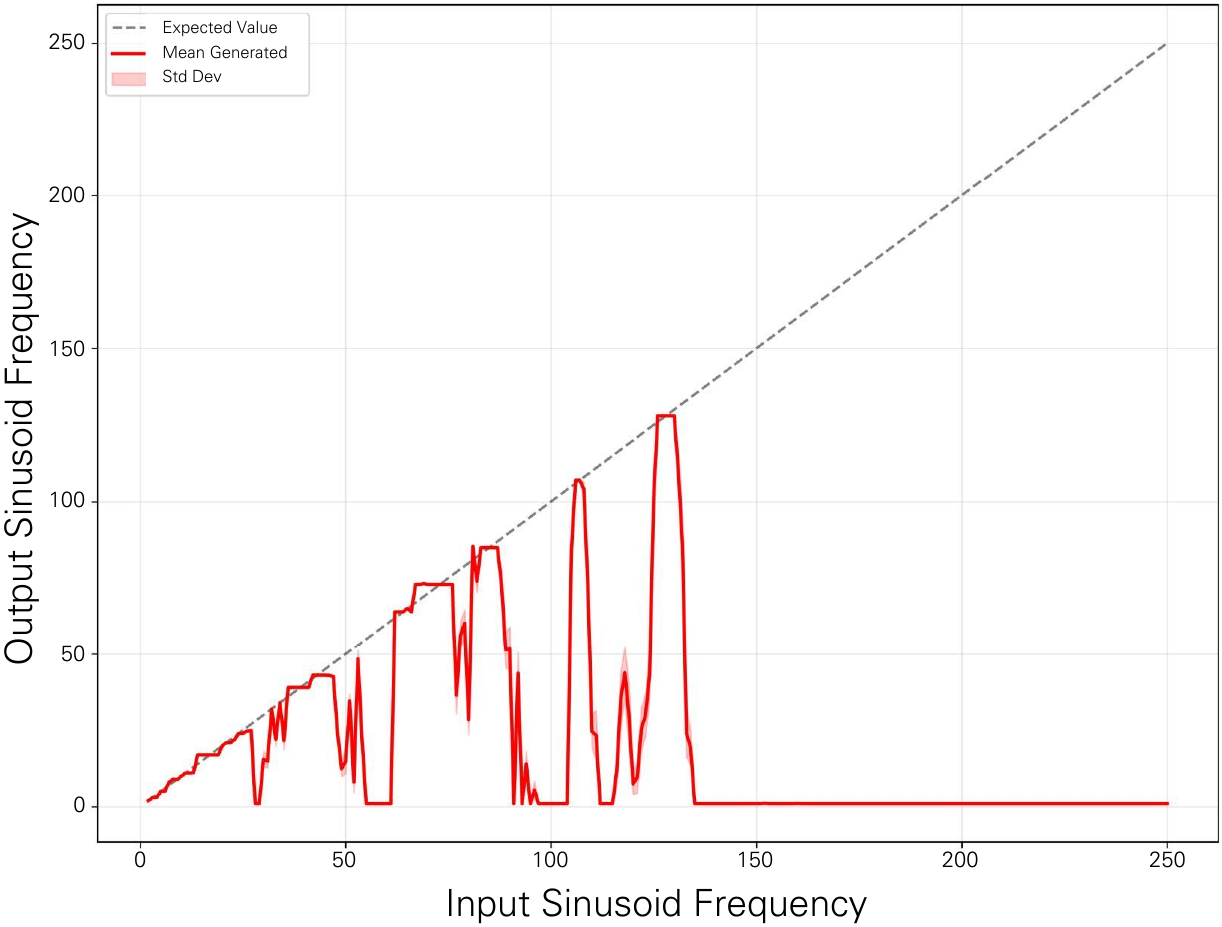}
    \caption{Comparison of input (x-axis) versus output (y-axis) sinusoid frequencies. The solid red line indicates the mean generated frequency, with the standard deviation shown by the light red shaded area. The dotted grey line represents the expected ideal values.}
    \label{fig:results:4}
\end{figure}

The baseline model yields an \gls{rmse} of 137.71, producing stable and accurate generations primarily up to ${\sim}25$ Hz. Beyond this threshold, the generation becomes increasingly unstable, culminating in a complete model collapse at frequencies above 130 Hz, where the output abruptly degenerates (as illustrated in Figure \ref{fig:results:4}).

\subsubsection{Single-layer interventions}

Layer-wise erasure shows heterogeneous effects. Removing the concept from Layer~0 reduces the global error (see \textit{Test 0} in \Cref{table_frequency_degr_merged}), indicating that early representations introduce interference. This phenomenon aligns with the ``Hydra Effect'' described by McGrath et al. \cite{McGrath2023TheHE}, a form of adaptive computation where ablating one component of a language model induces compensation in another. While the original study focused on attention layers, our findings suggest a generalized instance of this effect, wherein the ablation of specific representation vectors (via \gls{leace}) triggers similar compensatory mechanisms in downstream layers. 

Erasure in the intermediate blocks (Layers~1–3) increases \gls{rmse}, with Block~3 producing the strongest single-layer degradation. The output layer (4) behaves similarly to Layer~0 but with larger magnitude: its removal alone reduces \gls{rmse}, implying that the final projection may amplify spectral error rather than stabilise it.

\subsubsection{Cumulative interventions}

Joint erasure reveals strong inter-layer dependency. Although removing the output layer alone appears benign (\gls{rmse} $134.25$), it becomes critical when combined with deeper blocks. Configurations \texttt{01234} (\ie jointly intervening on all the layers), \texttt{1234} (\ie jointly intervening on all but the first layer), \texttt{234}, and \texttt{34} all produce substantial degradation, with \texttt{1234} reaching the highest \gls{rmse} ($140.75$), an increase of $3.04$ over baseline. 

Assuming complete loss of frequency information corresponds to predictions collapsing towards $f_i=0$ over the 2–250 Hz range, the theoretical upper bound is approximately $145.06$. The observed gap indicates residual information, likely encoded non-linearly or in attention components not affected by \gls{leace}, which acts linearly.

\section{Conclusion}
\label{sec:conclusions}
This study examined whether and how Amazon Chronos-Bolt encodes frequency information in its decoder representations under controlled sinusoidal inputs. Using hierarchical band discrimination tasks and online \gls{mdl} probes, we found that frequency information is strongly and linearly extractable across the four macro decoder blocks. Space Saving values approach unity and classification accuracy saturates, indicating near-complete compression relative to a uniform baseline. Control experiments with random labels yield negative Space Saving, confirming probe selectivity.

In contrast, the output projection layer exhibits reduced compression and lower accuracy, particularly at lower frequencies. This indicates that although frequency information is present in intermediate representations, its linear accessibility diminishes at the final projection stage. Frequency-resolved evaluation further reveals structured degradation patterns concentrated near binary decision thresholds and at specific harmonics, most notably multiples of the $32$ Hz patch frequency induced by the architectural configuration ($P=16$, $f_s=512$ Hz). When $f = k(f_s/P)$, successive patches become identical, producing token-level degeneracy and reduced effective temporal resolution. While patch-stride aliasing accounts for many observed failures, residual anomalies suggest additional interactions with positional encoding or non-linear transformations.

Intervention via sequential \gls{leace} demonstrates that erasing a coarse binary abstraction of frequency is sufficient to impair generation. The impact of erasure is heterogeneous across layers: intermediate blocks, especially deeper ones, produce the largest spectral degradation, whereas isolated removal of the earliest or final layer can yield compensatory effects. Cumulative interventions reveal substantial inter-layer redundancy. Even the most destructive configurations do not reach the theoretical upper bound of spectral error, indicating that frequency information is distributed beyond the linear feed-forward subspaces targeted by \gls{leace} and likely involves non-linear components.

Overall, frequency content in Chronos is robustly encoded in intermediate decoder states, partially attenuated at the output layer, sensitive to architectural synchronisation at specific harmonics, and redundantly distributed across layers. These findings clarify the internal treatment of a fundamental signal attribute and highlight practical regimes requiring care, particularly around patch-aligned frequencies. More broadly, the results illustrate how information-theoretic probing combined with causal intervention can provide mechanistic insight into temporal foundation models, supporting their informed use in signal processing and information fusion applications.

Future work should test whether these findings generalise to more complex and non-stationary signals, and examine encoder and cross-attention states to localise where frequency information becomes accessible. Further analysis of patch configuration and sampling rate may also clarify whether harmonic degradation effects are inherent to the architecture or can be mitigated through adaptation.

\section*{Acknowledgments}
The work was partially supported by the European Office of Aerospace Research \& Development under award numbers FA8655-22-1-7017 and FA8655-25-1-7067, and by the US DEVCOM Army Research Laboratory (ARL) under Cooperative Agreement \#W911NF2220243. Any opinions, findings, and conclusions or recommendations expressed in this material are those of the author(s) and do not necessarily reflect the views of the authors or of the United States government.
The authors used GPT-5.2 to improve readability and language. They reviewed and edited the content as needed, and they take full responsibility for the publication's content. Distribution unlimited: AFRL-2026-1317.

\bibliographystyle{IEEEtran} 
\bibliography{additional, references, biblio}

@misc{ansari2025chronos2univariateuniversalforecasting,
      title={Chronos-2: From Univariate to Universal Forecasting}, 
      author={Abdul Fatir Ansari and Oleksandr Shchur and Jaris Küken and Andreas Auer and Boran Han and Pedro Mercado and Syama Sundar Rangapuram and Huibin Shen and Lorenzo Stella and Xiyuan Zhang and Mononito Goswami and Shubham Kapoor and Danielle C. Maddix and Pablo Guerron and Tony Hu and Junming Yin and Nick Erickson and Prateek Mutalik Desai and Hao Wang and Huzefa Rangwala and George Karypis and Yuyang Wang and Michael Bohlke-Schneider},
      year={2025},
      eprint={2510.15821},
      archivePrefix={arXiv},
      primaryClass={cs.LG},
      url={https://arxiv.org/abs/2510.15821}, 
}

@article{10.5555/3455716.3455856,
author = {Raffel, Colin and Shazeer, Noam and Roberts, Adam and Lee, Katherine and Narang, Sharan and Matena, Michael and Zhou, Yanqi and Li, Wei and Liu, Peter J.},
title = {Exploring the limits of transfer learning with a unified text-to-text transformer},
year = {2020},
issue_date = {January 2020},
publisher = {JMLR.org},
volume = {21},
number = {1},
issn = {1532-4435},
journal = {J. Mach. Learn. Res.},
month = jan,
articleno = {140},
numpages = {67}
}

@article{raffel2020exploring,
  title={Exploring the limits of transfer learning with a unified text-to-text transformer},
  author={Raffel, Colin and Shazeer, Noam and Roberts, Adam and Lee, Katherine and Narang, Sharan and Matena, Michael and Zhou, Yanqi and Li, Wei and Liu, Peter J},
  journal={Journal of machine learning research},
  volume={21},
  number={140},
  pages={1--67},
  year={2020}
}

@inproceedings{hewitt-liang-2019-designing,
    title = "Designing and Interpreting Probes with Control Tasks",
    author = "Hewitt, John  and
      Liang, Percy",
    editor = "Inui, Kentaro  and
      Jiang, Jing  and
      Ng, Vincent  and
      Wan, Xiaojun",
    booktitle = "Proceedings of the 2019 Conference on Empirical Methods in Natural Language Processing and the 9th International Joint Conference on Natural Language Processing (EMNLP-IJCNLP)",
    month = nov,
    year = "2019",
    address = "Hong Kong, China",
    publisher = "Association for Computational Linguistics",
    url = "https://aclanthology.org/D19-1275/",
    doi = "10.18653/v1/D19-1275",
    pages = "2733--2743"
}

@inproceedings{voita-titov-2020-information,
    title = "Information-Theoretic Probing with Minimum Description Length",
    author = "Voita, Elena  and
      Titov, Ivan",
    editor = "Webber, Bonnie  and
      Cohn, Trevor  and
      He, Yulan  and
      Liu, Yang",
    booktitle = "Proceedings of the 2020 Conference on Empirical Methods in Natural Language Processing (EMNLP)",
    month = nov,
    year = "2020",
    address = "Online",
    publisher = "Association for Computational Linguistics",
    url = "https://aclanthology.org/2020.emnlp-main.14/",
    doi = "10.18653/v1/2020.emnlp-main.14",
    pages = "183--196"
}

@article{wang2025spectral,
  title={Spectral Predictability as a Fast Reliability Indicator for Time Series Forecasting Model Selection},
  author={Wang, Oliver and Quan, Pengrui and Yang, Kang and Srivastava, Mani},
  journal={arXiv preprint arXiv:2511.08884},
  year={2025}
}

@article{McGrath2023TheHE,
  title={The Hydra Effect: Emergent Self-repair in Language Model Computations},
  author={Tom McGrath and Matthew Rahtz and J{\'a}nos Kram{\'a}r and Vladimir Mikulik and Shane Legg},
  journal={ArXiv},
  year={2023},
  volume={abs/2307.15771},
  url={https://api.semanticscholar.org/CorpusID:260334719}
}

@inproceedings{10.5555/3666122.3669006,
author = {Belrose, Nora and Schneider-Joseph, David and Ravfogel, Shauli and Cotterell, Ryan and Raff, Edward and Biderman, Stella},
title = {LEACE: perfect linear concept erasure in closed form},
year = {2023},
publisher = {Curran Associates Inc.},
address = {Red Hook, NY, USA},
booktitle = {Proceedings of the 37th International Conference on Neural Information Processing Systems},
articleno = {2884},
numpages = {20},
location = {New Orleans, LA, USA},
series = {NIPS '23}
}

@misc{concepterasure_github,
  author = {Belrose, Nora},
  title = {Concept Erasure: {LEACE} implementation},
  year = {2023},
  publisher = {GitHub},
  journal = {GitHub repository},
  howpublished = {\url{https://github.com/EleutherAI/concept-erasure}}
}

@misc{chronos_github,
  author = {Amazon Science},
  title = {Chronos Forecasting},
  year = {2024},
  publisher = {GitHub},
  journal = {GitHub repository},
  howpublished = {\url{https://github.com/amazon-science/chronos-forecasting}},
}

@misc{ansari2024chronoslearninglanguagetime,
      title={Chronos: Learning the Language of Time Series}, 
      author={Abdul Fatir Ansari and Lorenzo Stella and Caner Turkmen and Xiyuan Zhang and Pedro Mercado and Huibin Shen and Oleksandr Shchur and Syama Sundar Rangapuram and Sebastian Pineda Arango and Shubham Kapoor and Jasper Zschiegner and Danielle C. Maddix and Hao Wang and Michael W. Mahoney and Kari Torkkola and Andrew Gordon Wilson and Michael Bohlke-Schneider and Yuyang Wang},
      year={2024},
      eprint={2403.07815},
      archivePrefix={arXiv},
      primaryClass={cs.LG},
      url={https://arxiv.org/abs/2403.07815}, 
}

@INPROCEEDINGS{11180285,
  author={Lessa, Sebastião Santos and Lucas, Alexandre},
  booktitle={2025 IEEE Kiel PowerTech}, 
  title={Synthetic Data Generation for Time Series Imputation: Comparing the Foundation Model Chronos with Established Methods}, 
  year={2025},
  volume={},
  number={},
  pages={1-6},
  doi={10.1109/PowerTech59965.2025.11180285}}

@misc{park2025unicastunifiedmultimodalprompting,
      title={UniCast: A Unified Multimodal Prompting Framework for Time Series Forecasting}, 
      author={Sehyuk Park and Soyeon Caren Han and Eduard Hovy},
      year={2025},
      eprint={2508.11954},
      archivePrefix={arXiv},
      primaryClass={cs.AI},
      url={https://arxiv.org/abs/2508.11954}, 
}

@misc{olivares2025realisticevaluationcrossfrequencytransfer,
      title={A More Realistic Evaluation of Cross-Frequency Transfer Learning and Foundation Forecasting Models}, 
      author={Kin G. Olivares and Malcolm Wolff and Tatiana Konstantinova and Shankar Ramasubramanian and Boris Oreshkin and Andrew Gordon Wilson and Andres Potapczynski and Willa Potosnak and Michael W. Mahoney and Mengfei Cao and Dmitry Efimov},
      year={2025},
      eprint={2509.19465},
      archivePrefix={arXiv},
      primaryClass={cs.LG},
      url={https://arxiv.org/abs/2509.19465}, 
}

@mastersthesis{PerakylaYear,
  author       = {Peräkylä, Luukas},
  title        = {FORECASTING HEART RATE VARIABILITY USING WEARABLE SENSOR DATA},
  school       = {Tampere University},
  year         = {2025},
  type         = {Bachelor’s thesis},
  url          = {https://trepo.tuni.fi/bitstream/handle/10024/231303/PerakylaLuukas.pdf}
}

@INPROCEEDINGS{10706320,
  author={Cominelli, Marco and Gringoli, Francesco and Kaplan, Lance M. and Srivastava, Mani B. and Bihl, Trevor and Blasch, Erik P. and Iyer, Nandini and Cerutti, Federico},
  booktitle={2024 27th International Conference on Information Fusion (FUSION)}, 
  title={Neuro-Symbolic Fusion of Wi-Fi Sensing Data for Passive Radar with Inter-Modal Knowledge Transfer}, 
  year={2024},
  volume={},
  number={},
  pages={1-8},
  doi={10.23919/FUSION59988.2024.10706320}}

@INPROCEEDINGS{11123929,
  author={Bresciani, Christian and Lavazza, Luca and Cominelli, Marco and Han, Liying and Dong, Gaofeng and Gringoli, Francesco and Kaplan, Lance M. and Srivastava, Mani B. and Bihl, Trevor and Blasch, Erik P. and Knutson, Felix J. and Cerutti, Federico},
  booktitle={2025 28th International Conference on Information Fusion (FUSION)}, 
  title={Preliminary Insights Into Resource-Constrained Neuro-Symbolic Causal Complex Event Processing}, 
  year={2025},
  volume={},
  number={},
  pages={1-8},
  doi={10.23919/FUSION65864.2025.11123929}}

@inproceedings{NEURIPS2019_bdbca288,
 author = {Paszke, Adam and Gross, Sam and Massa, Francisco and Lerer, Adam and Bradbury, James and Chanan, Gregory and Killeen, Trevor and Lin, Zeming and Gimelshein, Natalia and Antiga, Luca and Desmaison, Alban and Kopf, Andreas and Yang, Edward and DeVito, Zachary and Raison, Martin and Tejani, Alykhan and Chilamkurthy, Sasank and Steiner, Benoit and Fang, Lu and Bai, Junjie and Chintala, Soumith},
 booktitle = {Advances in Neural Information Processing Systems},
 editor = {H. Wallach and H. Larochelle and A. Beygelzimer and F. d\textquotesingle Alch\'{e}-Buc and E. Fox and R. Garnett},
 pages = {},
 publisher = {Curran Associates, Inc.},
 title = {PyTorch: An Imperative Style, High-Performance Deep Learning Library},
 url = {https://proceedings.neurips.cc/paper_files/paper/2019/file/bdbca288fee7f92f2bfa9f7012727740-Paper.pdf},
 volume = {32},
 year = {2019}
}

@article{IJISTECH242,
	author = {Surya Nasution and Imam Saputra},
	title = {Performance Analysis of Lossless Type of Compression Algorithm in Compression Data},
	journal = {IJISTECH (International Journal of Information System and Technology)},
	volume = {6},
	number = {3},
	year = {2022},
	issn = {2580-7250},	pages = {296--302},	doi = {10.30645/ijistech.v6i3.242},
	url = {https://ijistech.org/ijistech/index.php/ijistech/article/view/242}
}

@inproceedings{10.1145/3292500.3330701,
author = {Akiba, Takuya and Sano, Shotaro and Yanase, Toshihiko and Ohta, Takeru and Koyama, Masanori},
title = {Optuna: A Next-generation Hyperparameter Optimization Framework},
year = {2019},
isbn = {9781450362016},
publisher = {Association for Computing Machinery},
address = {New York, NY, USA},
url = {https://doi.org/10.1145/3292500.3330701},
doi = {10.1145/3292500.3330701},
booktitle = {Proceedings of the 25th ACM SIGKDD International Conference on Knowledge Discovery \& Data Mining},
pages = {2623–2631},
numpages = {9},
location = {Anchorage, AK, USA},
series = {KDD '19}
}

@inproceedings{
bornschein2023sequential,
title={Sequential Learning of Neural Networks for Prequential {MDL}},
author={Jorg Bornschein and Yazhe Li and Marcus Hutter},
booktitle={The Eleventh International Conference on Learning Representations },
year={2023},
url={https://openreview.net/forum?id=dMMPUvNSYJr}
}

\end{document}